\documentclass[11pt]{article}
\usepackage{acl}

\usepackage{times}
\usepackage{latexsym}
\usepackage[T1]{fontenc}
\usepackage[utf8]{inputenc}
\usepackage{microtype}
\usepackage{graphicx}
\usepackage{booktabs}
\usepackage{amsmath}
\usepackage{amssymb}
\usepackage{xcolor}
\usepackage{url}
\usepackage{tikz}
\usetikzlibrary{arrows.meta,positioning,fit,backgrounds,calc}

\graphicspath{{figures/}}

\definecolor{cBlue}{HTML}{0F4D92}
\definecolor{cGreen}{HTML}{8BCF8B}
\definecolor{cRed}{HTML}{B64342}
\definecolor{cNeutral}{HTML}{CFCECE}
\definecolor{cTeal}{HTML}{42949E}
\definecolor{cViolet}{HTML}{9A4D8E}

\newcommand{\coderepo}{ will be released.}

\title{Tool-Call Dependency Structure is Linearly Decodable in LLM Agent Residual Streams}
\author{Tianda Sun \and Dimitar Kazakov \\
        University of York \\
        Department of Computer Science \\
        Heslington, York \\
        YO10 5DD \\
        \texttt{tianda.sun@york.ac.uk} \\
        \texttt{dimitar.kazakov@york.ac.uk}}

\begin{document}
\maketitle

% ---------- abstract ----------
% Abstract — result-first (Reviewer P0-2): lead with the finding, ONE headline
% number, causal as co-headline, method machinery moved to the body.
\begin{abstract}
Tool-using LLM agents produce trajectories whose calls form a directed dependency graph: earlier tool outputs supply arguments to later calls. Whether this execution structure is represented inside the model is unknown; prior structural probes have targeted static code or chain-of-thought text, not an agent's run-time call graph. A low-capacity edge probe on the residual stream of Qwen3-32B decodes the tool-call dependency graph well above both a Hewitt--Liang random-label control and a positional baseline. A counterfactual contrast between value corruption and structural perturbation indicates the signal tracks abstract topology rather than identifier values, and replicates under an independent, non-substring oracle. The non-positional component replicates on three further interactive multi-hop benchmarks and attenuates as call order alone becomes a sufficient proxy for dependency, vanishing in single-shot planning. Per-layer activation patching shifts the probe at a later, non-patched boundary, evidence that the representation propagates rather than passively reads out, though the realised tool call does not move. To our knowledge this is the first structural probe of an LLM agent's runtime tool-call dependency graph. Our claims concern representation, not behavioural control, and span two model families and one primary domain.
\end{abstract}

% ---------- overview Figure 1 (full-width float, lands top of p1/p2) ----------
% Figure 1 — MONEY FIGURE (single-column, page 1). Panel A: what the object
% is (trajectory -> oracle DAG over the residual stream -> edge probe).
% Panel B: the finding (decodable AND causal, by layer) = F0_money.pdf.
% Numbers: ../../DATA_INVENTORY.md only. Caption is self-contained.
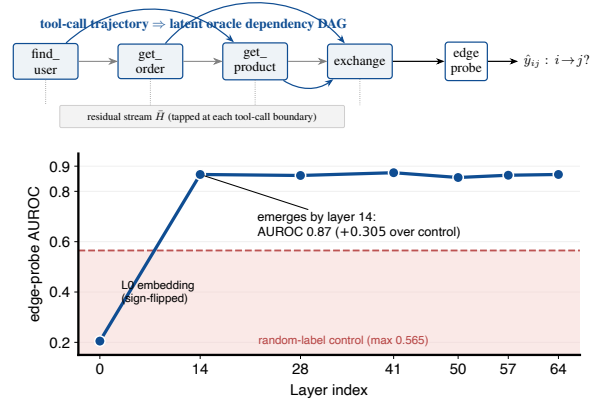
\begin{figure}[t]
\centering
% --- Panel A: the object of study (compact TikZ schematic) ---
\resizebox{\columnwidth}{!}{%
\begin{tikzpicture}[
  font=\sffamily,
  >={Stealth[length=2mm]},
  call/.style={draw=cBlue,thick,rounded corners=2pt,fill=cBlue!7,
               minimum width=15mm,minimum height=8mm,align=center,font=\small},
  dep/.style={->,cBlue,thick},
  seq/.style={->,gray,thin},
]
\node[call] (c0) {find\_\\user};
\node[call,right=9mm of c0]  (c1) {get\_\\order};
\node[call,right=9mm of c1]  (c2) {get\_\\product};
\node[call,right=9mm of c2]  (c3) {exchange};
\foreach \a/\b in {c0/c1,c1/c2,c2/c3}{\draw[seq] (\a)--(\b);}
\draw[dep,bend left=42]  (c0) to (c2);
\draw[dep,bend left=50]  (c1) to (c3);
\draw[dep,bend right=34] (c2) to (c3);
% residual stream strip
\node[draw=cNeutral,fill=cNeutral!25,minimum width=66mm,minimum height=5mm,
      rounded corners=1pt,below=6mm of c1.south,xshift=12mm,font=\scriptsize]
      (rs) {residual stream $\bar H$ (tapped at each tool-call boundary)};
\foreach \c in {c0,c1,c2,c3}{\draw[gray,thin,densely dotted] (\c.south)--(\c.south|-rs.north);}
% probe
\node[draw=cBlue,thick,rounded corners=2pt,fill=white,right=12mm of c3,
      minimum height=8mm,font=\small,align=center] (pr)
      {edge\\probe};
\draw[->,thick] (c3.east) -- (pr.west);
\node[right=7mm of pr,font=\small] (y) {$\hat y_{ij}:\, i\!\to\! j$?};
\draw[->,thick] (pr) -- (y);
\node[above=1.5mm of c1.north,xshift=10mm,font=\small\bfseries,color=cBlue]
      {tool-call trajectory $\Rightarrow$ latent oracle dependency DAG};
\end{tikzpicture}}

\vspace{3pt}
% --- Panel B: the finding (decodable AND causal, by layer) ---
\includegraphics[width=\columnwidth]{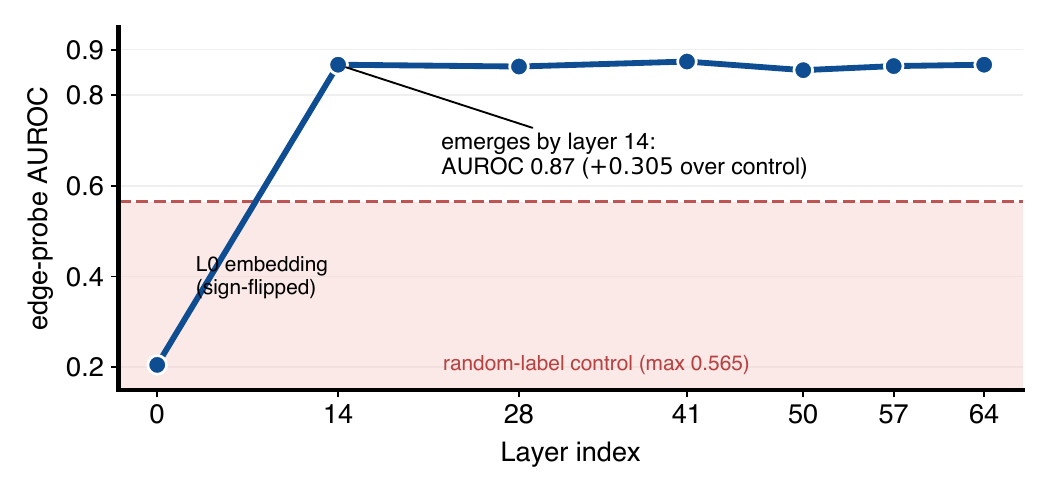}

\caption{\textbf{The tool-call dependency graph is linearly decodable from an LLM agent's residual stream.} \textbf{Top:} tool-using LLM agents call functions sequentially, and earlier outputs supply arguments to later calls, inducing a latent dependency DAG (an edge $i\!\to\!j$ iff call $i$'s output supplies an argument of call $j$); a low-capacity logistic probe reads each edge from the frozen residual stream. \textbf{Bottom:} the edge becomes linearly decodable by layer 14, well above a Hewitt--Liang random-label control. Activation patching at the source call shifts the probe's prediction toward a donor trajectory's structure at every layer $\geq\!14$ (\S\ref{sec:causal}, Appendix~\ref{app:xbench}); the non-positional component generalises across interactive multi-hop benchmarks (\S\ref{sec:claimB}).}
\label{fig:overview}
\end{figure}

% ---------- body (≤ 8 pages, ARR long) — conventional ACL/EMNLP arc ----------
% Introduction — long paper §1. Paras 1-2 adapted from the FROZEN style-audited
% short-paper §1; para 3 rewritten for the long paper's 3 contributions
% (decodable+causal / generalises+modulated / benchmark grid). Numbers: DATA_INVENTORY §1.
\section{Introduction}
\label{sec:intro}

Mechanistic interpretability has yielded sharp characterisations of how large language models represent linguistic and propositional structure: structural probes recover syntactic dependency trees from contextual embeddings \citep{hewitt2019structural}, board-state probes reveal compact world models in game-trained networks \citep{li2023othello, yuan2025othello}, and dissection methods locate factual knowledge in middle-layer feed-forward modules \citep{geva2023dissecting, heinzerling2024monotonic}. These results share a methodological core: train a small probe to read an abstract structure from frozen activations, and use control tasks \citep{hewitt2019control} and checks for interpretability illusions \citep{friedman2023illusions} to verify that the probe recovers structure rather than memorising labels. Recent work has extended structural probing to numerical structure \citep{levy2025numbers, elshangiti2025geometry}, propositional commitments \citep{feng2025propositional}, spatial world models \citep{li2026spatial}, and planning in chain-of-thought traces \citep{dong2025planning, zhang2025reasoning}.

However, the structural-probe literature has not engaged with a setting now central to deployed LLM systems: tool-using agents that interleave model generation with external function calls \citep{schick2023toolformer, yao2023react, yao2024taubench}. Agent trajectories carry a non-trivial structural object --- the directed acyclic graph (DAG) of tool-call dependencies, in which an edge $i \to j$ records that the output of call $i$ supplies an argument of call $j$. \emph{Whether this dependency graph is encoded in the agent's residual stream, whether that encoding is causal, and whether it is a property of agents in general or of one benchmark, are open}, despite emerging interest in agent-state interpretability for behavioural drift \citep{wu2026agentdrift}, retrieval auditing \citep{healy2026agenttool}, and deception detection \citep{abdelnabi2024tasktracker, apollo2025deception}.

We address this gap with a Hewitt--Liang-style logistic-regression edge probe on the residual stream of Qwen3-32B \citep{qwen3_2025}, conditioned on a positional baseline \citep{hewitt2021conditional} so the signal is non-trivial (Figure~\ref{fig:overview}). We make three contributions.

\textbf{(1) The tool-call dependency graph is linearly decodable} (\S\ref{sec:claimA}): the inter-call dependency DAG of LLM-agent tool trajectories is linearly readable from frozen residual streams, above random-label, positional, random-init, surface-form, and probe-capacity controls.

\textbf{(2) Topology sensitivity} (\S\ref{sec:claimA}): a counterfactual contrast indicates the probe tracks abstract dependency topology rather than identifier values; as supporting evidence that the representation is not a passive readout, per-layer activation patching shifts a later, non-patched probe readout, though the agent's realised tool choice does not move.

\textbf{(3) Cross-domain modulation} (\S\ref{sec:claimB}): the non-positional contribution follows a falsifiable prediction --- positive on interactive multi-hop benchmarks \citep{toolhop, complexfuncbench}, vanishing where call order alone suffices \citep{taskbench} or transitive structure is absent \citep{bfcl} --- and it reproduces on independently provided ground-truth DAGs \citep{taskbench}.

\S\ref{sec:claimA} develops the first two contributions; \S\ref{sec:claimB} develops the third.
   % §1 Introduction
% Related Work — long paper §2 (NEW dedicated early section; conventional
% ACL/EMNLP arc, ReDeEP-style §2). Content lifted from the old combined
% Discussion+Related §8. No numbers.
\section{Related Work}
\label{sec:related}

\textbf{Structural probes.} The original Hewitt--Manning probe \citep{hewitt2019structural} read syntactic trees from BERT; subsequent work has read board states \citep{li2023othello, yuan2025othello}, propositional commitments \citep{feng2025propositional, heinzerling2024monotonic}, numeric structure \citep{levy2025numbers}, and spatial coordinates \citep{li2026spatial} from frozen activations. The methodological core is shared: a low-capacity probe reads off an abstract structure, validated with control tasks \citep{hewitt2019control}, conditional probing \citep{hewitt2021conditional}, and checks for interpretability illusions \citep{friedman2023illusions}. We extend this line by treating an agent's tool-call DAG --- and its transitive closure --- as the probe target. The closest precedents probe \emph{graph} structure --- reasoning-DAG adjacency from hidden states \citep{zhong2026chainsdags}, static code data-flow edges \citep{troshin2022probingcode, guo2021graphcodebert}, and semantic-relation graphs decoded from text \citep{polar2026} --- or scalar CoT plans \citep{dong2025planning, zhang2025reasoning, elshangiti2025geometry}; we differ in that our edges are \emph{side-effecting tool calls} with an independently constructed ground-truth DAG (not tokens, static code, or text-derived semantic relations), and we add activation patching and a topology-vs-value dissociation these probe-only studies lack.

\textbf{Agent interpretability.} Recent work has examined retrieval-head identification \citep{wu2025retrievalhead}, tool-output tracking \citep{abdelnabi2024tasktracker, healy2026agenttool}, deception detection \citep{apollo2025deception}, behavioural drift \citep{wu2026agentdrift}, hallucination editing \citep{sun2025redeep}, and knowledge conflicts \citep{zhao2025knowledge}. Closest to our setting, \citet{tatsat2026blackbox} probe agent residual streams for \emph{single-step} tool-need and tool-risk (with feature ablation for a causal effect on tool selection), \citet{sun2026already} decode a per-step tool-necessity signal from hidden states, and \citet{wu2026toolcalling, wang2026asa} read and steer the chosen tool's \emph{identity} or intent from mid-layer activations. These decode \emph{per-step} quantities (whether to call a tool, or which one); we instead decode the \emph{relational} structure linking calls --- the inter-call dependency edge $(i,j)$ --- with per-layer activation patching of the dependency representation plus a topology-vs-value dissociation. The two approaches target complementary objectives: per-step decisions versus multi-step structure. To our knowledge this is the first structural probe of an LLM agent's tool-call execution graph at the inter-call dependency level.

\textbf{Tool-use benchmarks and the interactive$\times$multi-hop axis.} Agent tool-use benchmarks vary along two axes that turn out to matter for our analysis: whether tool calls are issued \emph{interactively} with execution feedback or emitted in a single shot, and whether tasks require \emph{multi-hop} chained dependencies or only direct ones \citep{yao2024taubench, schick2023toolformer, yao2023react}. The conjunction we rely on --- benchmarks that are \emph{both} interactive and multi-hop --- is, by our own benchmark sweep, sparsely instantiated; we use this axis both to select evaluation benchmarks and as an explanatory variable for where our effect is measurable.
   % §2 Related Work (dedicated, early)
% Method — long paper §3 (was "Setup"; renamed for conventional ACL/EMNLP arc).
% \label kept as sec:setup so all existing \ref{sec:setup} resolve unchanged.
% Numbers: ../../DATA_INVENTORY.md §2 (+ §3.1 pair counts).
\section{Method}
\label{sec:setup}

We run Qwen3-32B \citep{qwen3_2025} as the policy of an agent in the retail split of $\tau$-bench \citep{yao2024taubench}, a multi-turn customer-service benchmark in which the agent answers user requests by issuing typed JSON tool calls against a simulated retail backend. Retail is our \emph{primary} domain; \S\ref{sec:crossdomain} re-runs the identical pipeline, unchanged, on four further agent benchmarks to test cross-domain generality. We log the raw residual stream of all 65 representation layers (token embedding plus 64 transformer blocks) at every assistant decode boundary.

\paragraph{Oracle tool-call DAG.} For each trajectory containing $\geq 2$ tool calls we construct the \emph{oracle tool-call DAG} (the binary edge label our probe will learn to predict): a directed edge $i \to j$ is added when a contiguous, case-sensitive, whitespace-normalised character substring of length $\geq 4$ of call $i$'s output occurs verbatim in the serialised JSON arguments of call $j$, i.e.\ call $i$'s result supplies an argument of call $j$ --- for example, if \texttt{get\_user} returns a field \texttt{user\_id=123} and a later \texttt{get\_order} call includes \texttt{user\_id=123} in its arguments, we add the edge \texttt{get\_user}~$\to$~\texttt{get\_order} (exact normalisation and matching procedure in Appendix~\ref{app:impl}; oracle code released). This length-$\geq 4$ substring rule trades recall for precision; \S\ref{sec:crossdomain} shows the central finding also holds against an \emph{independently constructed} ground-truth DAG (TaskBench), so it is not an artefact of the substring heuristic. Because the oracle uses textual overlap, \S\ref{sec:pairwise} additionally pairs the residual probe with a non-residual surface-form decoder as a textual-overlap ceiling control. On $\tau$-bench retail this yields $1{,}129$ ordered pairs $(i,j)$ with $i<j$ across $105$ trajectories ($286$ positive edges, $843$ negatives); the probe target is binary edge presence. We additionally define the \emph{transitive-only} label set (\S\ref{sec:chain}): $y{=}1$ iff $i$ is a multi-hop ancestor of $j$ in the oracle DAG and $(i,j)$ is \emph{not} a direct edge.

\paragraph{Probe.} We train an $L_2$-regularised logistic-regression probe ($C{=}0.01$, balanced class weights; \texttt{StandardScaler} fit on the training fold only; full hyper-parameters in Appendix~\ref{app:impl}) on the concatenation $x_{ij}=[\bar{H}_i\,;\,\bar{H}_j]$ of residual-stream vectors at the terminating boundary token of calls $i$ and $j$, each mean-pooled across a fixed 7-layer set $\{0,14,28,41,50,57,64\}$ (final dimension $10{,}240$; the canonical \textbf{V1} configuration of Appendix~\ref{app:impl}). A linear probe of this capacity is a deliberately low-capacity reader \citep{hewitt2019control,nordby2026probescaling}: it can recover structure already linearly present in the residual stream but cannot synthesise it.

\paragraph{Evaluation.} We use per-trajectory leave-one-group-out cross-validation (\textsc{logo}; 105 folds on retail) so that no token from a held-out trajectory enters its own training fold; per-boundary CV leaks within-trajectory context and is excluded. We report AUROC over all out-of-fold predictions and a 95\% BCa bootstrap CI from $2{,}000$ trajectory-level resamples. As the Hewitt--Liang random-label control \citep{hewitt2019control} we re-run the full pipeline with edge labels uniformly permuted ($500$ permutations for the main probe; $200$ for downstream variants) and report the empirical $p$-value. To separate structural encoding from a within-trajectory positional prior --- a probe could in principle predict edges from call ordering alone --- we use conditional probing \citep{hewitt2021conditional}: a 5-feature positional baseline $B=[i,j,j{-}i,n_{\mathrm{agent}},j/n_{\mathrm{agent}}]$, with the residual probe's AUROC \emph{gap} over $B$ as the contribution we report (paired-bootstrap CI). For cross-benchmark evaluation (\S\ref{sec:crossdomain}) resampling and \textsc{logo} are by \emph{task} identity, effective $N$ being the distinct-task count. Probe weights, control distributions, and bootstrap raw data\coderepo
         % §3 Method (was "Setup"; \label sec:setup kept)
% Results — long paper §4 (umbrella; two-claim grouping per confirmed decision).
% \input order: §4.1 Claim A (pairwise + topology-not-value + causal),
%               §4.2 Claim B (chain composition + cross-domain generality).
\section{Results and Discussion}
\label{sec:results}

\S\ref{sec:claimA} establishes the primary result --- the tool-call dependency graph is linearly decodable from the Qwen3-32B residual stream under a full control battery (Table~\ref{tab:main}) --- and adds \emph{supporting} evidence from activation patching that the signal propagates rather than passively reads out. \S\ref{sec:claimB} shows the non-positional component generalises across three further interactive multi-hop benchmarks and attenuates where call order alone predicts the graph. We mark the evidential strength of each claim where it is made: decodability and its transitive extension are established under controls; the cross-domain attenuation is a descriptive tendency, not a fitted law; and propagation is representational, not behavioural.

\subsection{Decoding the dependency graph}
\label{sec:claimA}

% Results 4.1.1 Pairwise dependency decoding (was top-level §3).
% Demoted to \subsubsection; \label{sec:pairwise} preserved for cross-refs.
% Numbers: ../../DATA_INVENTORY.md §3. Table 1 + F1 + F2 live here.
\subsubsection{Pairwise dependency decoding}
\label{sec:pairwise}

We first establish that the residual stream linearly encodes the \emph{pairwise} tool-call dependency relation; \S\ref{sec:chain} extends this to multi-hop composition. The controls below rule out the obvious alternatives --- random-label learnability, positional templates, surface-form textual overlap, and probe-capacity effects (Table~\ref{tab:main}).

\begin{table*}[t]
\centering
\caption{\textbf{The tool-call dependency edge is linearly decodable: AUROC 0.869, $+0.304$ over a random-label control and $+0.0775$ over a positional baseline.} \emph{Block 1 (information-theoretic null)}: Hewitt--Liang random-label permutation. \emph{Block 2 (structural-DAG null)}: decoded DAG vs scrambled / other-task oracle ($d$). \emph{Block 3 (representational alternative: positional prior, scaffold/template baseline, random-init, surface-form ceiling)}: conditional probing rules each out. \emph{Block 4 (model-capacity null)}: linear vs MLP, same features. Headline probe (row 1) is V1 with input $[\bar{H}_i,\bar{H}_j]$ (10{,}240-dim, the canonical configuration of \S\ref{sec:setup}); V0 adds a diff term (15{,}360-dim, AUROC 0.870; Appendix~\ref{app:impl}). All AUROCs use per-trajectory leave-one-group-out CV on the same $1{,}129$ pairs across 105 trajectories. The Reference~$\Delta$ column reports each row's gap to the natural control; CIs are 95\% paired-bootstrap (2000 resamples). Bold marks the headline number per block; full statistical detail in \S\ref{sec:pairwise}.}
\label{tab:main}
\small
\setlength{\tabcolsep}{4pt}
\resizebox{\textwidth}{!}{%
\begin{tabular}{l l r l}
\toprule
& \textbf{Metric} & \textbf{Value} & \textbf{Reference $\Delta$} \\
\midrule
\multicolumn{4}{l}{\emph{Control (AUROC on 1{,}129 pairs, per-trajectory leave-one-group-out CV):}} \\
\quad Edge probe, multi-layer V1 (\textbf{ours})               & AUROC       & \textbf{0.869} & --- \\
\quad Hewitt--Liang random-label control: mean (500 perms)        & AUROC       & 0.491 & $-0.378$ vs ours \\
\quad Hewitt--Liang random-label control: maximum                 & AUROC       & 0.565 & $-0.304$ vs ours \\
\midrule
\multicolumn{4}{l}{\emph{Structural baselines (median-SD shift, decoded vs alternative oracle):}} \\
\quad vs structurally-scrambled oracle (\textbf{primary})              & Cohen's $d$ & 0.84  & $p \approx 3{\times}10^{-12}$ \\
\quad vs random other-task oracle                                      & Cohen's $d$ & 0.19  & $p < 10^{-4}$ \\
\midrule
\multicolumn{4}{l}{\emph{Conditional probing \citep{hewitt2021conditional} (AUROC, same CV protocol):}} \\
\quad Position-only baseline $[i, j, j{-}i, n, j/n]$ (5 scalars, no residuals)  & AUROC       & 0.792 & template prior \\
\quad Trained probe + position (conditional probe, \textbf{ours})       & AUROC       & 0.869 & \textbf{+0.0775} over pos, CI $[+0.032,+0.127]$ \\
\quad Random-init Qwen3-32B + position                                  & AUROC       & 0.738 & $-0.053$ over pos, CI $[-0.101,-0.007]$ \\
\quad Surface-form decoder ($n$-gram + tool-name + pos, 36-dim)         & AUROC       & \textbf{0.830} & $\Delta{=}{+}0.039$, $[+0.012,+0.063]$ \\
\quad Scaffold baseline (tool-name $i/j$ + bigram + dist, no residuals)  & AUROC       & 0.823 & residual adds \textbf{+0.0413}, $[+0.019,+0.062]$ \\
\midrule
\multicolumn{4}{l}{\emph{Probe-architecture control (AUROC, same V1 features):}} \\
\quad MLP probe ($10{,}240 \to 1024 \to 256 \to 1$)                     & AUROC       & 0.866 & $-0.003$ vs linear V1 \\
\bottomrule
\end{tabular}}
\end{table*}

\textbf{The probe recovers tool-call edges far above chance.} The multi-layer V1 edge probe attains AUROC 0.869 over $1{,}129$ held-out pairs (95\% bootstrap CI $[0.801, 0.930]$; V0 reaches 0.870, Appendix~\ref{app:impl}). No random-label permutation of 500 \citep{hewitt2019control} exceeds it ($p<10^{-4}$, Table~\ref{tab:main}), so the layer-pooled features carry edge information not recoverable from a model fit to random labels \citep{friedman2023illusions}.

\textbf{The signal emerges in mid-stack layers.} The per-layer profile (Figure~\ref{fig:overview}, bottom; full control-band detail in Appendix~\ref{app:controls}) is computed from seven individually-trained single-layer probes. The embedding layer (L0) attains AUROC 0.205; since an AUROC below 0.5 is a sign-inverted boundary, this is 0.795 once flipped, so L0 already carries edge information that the \textsc{logo} probe happens to fit with a flipped sign (a random-init L0 probe sits at chance, 0.533; Appendix~\ref{app:controls}). AUROC then climbs sharply to 0.864 by layer 14 ($\sim$22\% depth), reaches a maximum of 0.875 at layer 41, and remains in $[0.855, 0.867]$ through the final layer 64. This pattern matches reports of mid-stack semantic emergence \citep{belrose2023tunedlens, geva2023dissecting} and rules out a representation that is purely output-routing.

\textbf{Decoded DAGs match the correct oracle more closely than structural shuffles.} Against a structurally-scrambled oracle, the decoded DAG's median edge-set symmetric difference (SD) shifts from 1.0 to 2.0 (Cohen's $d{=}0.84$, paired Wilcoxon $p \approx 3{\times}10^{-12}$); against a random other-task oracle, the shift is smaller ($d{=}0.19$, $p<10^{-4}$), reflecting $\tau$-bench retail's shared task template. Both confirm trajectory-specific dependency structure.

\textbf{The residual adds signal beyond position.} A 5-feature positional baseline $[i, j, j-i, n_{\mathrm{agent}}, j/n_{\mathrm{agent}}]$ attains AUROC 0.792 under the same CV. Following \citet{hewitt2021conditional}, a probe on $[\bar{H}_i, \bar{H}_j; i, j, j-i, n, j/n]$ attains AUROC 0.869, a conditional contribution of $\mathbf{+0.0775}$ over the positional baseline (95\% CI $[+0.032, +0.127]$; Table~\ref{tab:main}). A non-linear MLP on the same 5 positional features attains $0.781$, indistinguishable from the linear baseline. The same probe on random-init Qwen3-32B residuals contributes $-0.053$ \citep{heap2025randominit}, so the training-attributable gap, controlled for position, is $\mathbf{+0.131}$ AUROC (non-overlapping paired CIs). This $+0.0775$ pairwise contribution is the conservative baseline that \S\ref{sec:chain} more than doubles on the transitive task.

\textbf{The signal survives a stronger scaffold/template baseline.} Augmenting the positional scalars with tool-name identity of calls $i$ and $j$, their tool-name bigram, and call-distance yields a stronger template prior (AUROC 0.823 vs 0.792). The residual probe still contributes $\mathbf{+0.0413}$ over this scaffold baseline (95\% CI $[+0.019, +0.062]$, $P(\Delta{\leq}0){=}0$), and within matched tool-name-pair strata (23 strata) separates edges at within-stratum AUROC 0.774. The margin is smaller than the $+0.0775$ over position --- expected, as the scaffold prior is stronger --- but positive and significant: decodability is not a scaffold-template artefact.

\textbf{The residual probe beats a surface-form ceiling.} A non-residual decoder on $n$-gram overlap, tool-name features, and the 5 positional scalars (36-dim) attains AUROC 0.830; the residual probe exceeds this by paired $\Delta{=}+0.039$ (Table~\ref{tab:main}). Adding surface features on top of the residual probe gains only $+0.001$, suggesting the residual representation predictively subsumes the tested 36-dim surface feature set (Appendix~\ref{app:controls}).

\textbf{Input ablations.} Across five feature variants (V0--V4), V0 and V1 attain AUROC $\geq 0.869$; restricting to a single endpoint or layer drops AUROC by $0.03$--$0.08$, so we adopt V1 as canonical (Appendix~\ref{app:impl}).

% [F2 decoded-DAG-examples figure relocated to Appendix B (app:valuecorr) per
%  figure-strategy decision 2026-05-18; \label{fig:dags} now lives there.]

% Results 4.1.2 Topology, not value (was top-level §4 "Topology vs Value").
% Demoted to \subsubsection; \label{sec:dissoc} preserved. Numbers: DATA_INVENTORY §4.
\subsubsection{Topology-sensitive rather than value-specific}
\label{sec:dissoc}

Does the probe encode the abstract dependency \emph{topology}, or the specific identifier values flowing between calls? We compare its response to structural perturbations against matched single-field value corruptions.

\textbf{Decoded DAGs match the projected oracle.} Figure~\ref{fig:dags} shows four trajectories with $\geq 2$ tool calls; for each, we display the oracle DAG and the DAG decoded by thresholding the edge probe at its in-sample F1-optimal cut. On the agent's pair space (oracle projected to the agent's $n$ calls), per-panel SD is 0--2. The clean-condition whole-dataset distribution ($N{=}105$) has median SD~$=1.0$ (Appendix~\ref{app:valuecorr}). F1-thresholded decoded graphs are acyclic on all $208$ trajectories tested (105 clean $+$ 103 corrupted), so no post-hoc cycle resolution is required.

\textbf{Value corruption produces no detectable shift in the decoded structure.} We apply a counterfactual at the median tool-call index of each trajectory: swap a single \texttt{\_id} field in the call's output with another valid id and rerun the policy. On $N{=}103$ paired clean/corrupted trajectories the per-trajectory decoded-vs-oracle SD is statistically indistinguishable: median $\mathrm{SD}{=}1.0$ in both, mean shift $\Delta{=}{+}0.155$ edges, paired Wilcoxon $p{=}0.11$, $d{=}0.06$, drift AUROC 0.554 (Appendix~\ref{app:valuecorr}). Effective propagation $N \approx 32$: only 32 of the 120 corruptions land on a field a downstream oracle call would reference, so the non-rejection is underpowered against small effects; we interpret the $d{=}0.06$ vs $d{=}0.87$ contrast (paired CIs non-overlapping) as the load-bearing dissociation evidence, not the standalone null. The $\tau$-bench reward direction (clean $12.5\%$, corrupted $18.3\%$) is not statistically significant (Fisher exact $p{\approx}0.27$).

\textbf{A structural positive control calibrates the value-invariance null.} A structural perturbation that rewrites a single tool response to an empty result forces the agent to adapt its downstream plan (Appendix~\ref{app:valuecorr}). The decoded DAG on the intersected pair space differs from clean by median SD 1.0 (Wilcoxon $p=1.18\times10^{-11}$, $d=0.87$, $n=103$; 53\% non-zero shift): the probe IS responsive to structural drift. Decoded-vs-oracle SD is unchanged in expectation ($d=0.00$) because the task and its oracle DAG are unchanged by a response rewrite. \emph{The dissociation evidence is the contrast itself}: a near-zero effect for value corruption ($d{=}0.06$, approximate 95\% CI $[-0.13, +0.25]$) against a large effect for structural perturbation ($d{=}0.87$) on the same probe, the same pair space, and the same single-tool-call locality, though the two perturbations differ in magnitude. Whether a high-magnitude \emph{value} perturbation would also leave the probe invariant is untested (Limitations).

\textbf{The dissociation replicates on a second domain.} This test must hold the downstream plan fixed --- under free generation a value swap can itself trigger re-planning, which would move the decoded graph for a genuine structural reason --- so we evaluate it under teacher-forcing. On ComplexFuncBench (185 paired golden-trajectory replays) the contrast holds: the decoded DAG is invariant to a same-type value swap ($d{=}{+}0.018$, 95\% CI $[-0.13, +0.15]$, including zero) yet strongly responsive to an emptied observation ($d{=}{+}0.702$, CI $[+0.58, +0.83]$; Wilcoxon $p{=}1.5{\times}10^{-16}$). The topology-vs-value dissociation is thus cross-domain, not specific to $\tau$-bench retail.

\textbf{An independent oracle reproduces the finding.} A schema-typed value-equality oracle (no substring matching) agrees with the substring oracle at precision $1.0$, as a strict subset that \emph{excludes} exactly the substring-only edges most likely to be lexical. On this cleaner edge set decodability is itself high (direct AUROC $0.947$) and the conditional contribution over position \emph{rises} to $\mathbf{+0.289}$ ($P(\Delta{\leq}0){=}0$); the per-layer activation patching also replicates (Appendix~\ref{app:valuecorr}).

% Results 4.1.3 Causal evidence (causal half of old §6, demoted).
% \label{sec:causal} (new). F5 lives here. Numbers: DATA_INVENTORY §6.1.
\subsubsection{Representational propagation}
\label{sec:causal}

Decodability is correlational. Is the decoded structure load-bearing, or a passive readout? Per-layer activation patching offers \emph{supporting} evidence for propagation: a source-call patch shifts the dependency readout, and the shift survives to a later, non-patched boundary (App~\ref{app:xbench}).

% [F5 causal-patching figure relocated to Appendix E (app:xbench);
%  T40 forward-hook per-layer table added there 2026-05-20.]

Across 80 minimal pairs differing in one oracle edge (Figure~\ref{fig:causal}, App~\ref{app:xbench}), the feature-level patch is positive at every layer past the first fifth of the stack (peak $\Delta_{\mathrm{patch}}{=}{+}0.016$ at L57; L0 exactly zero), but is partly definitional since the patched layer feeds the probe. Read at a strictly later, \emph{non-patched} call-$j$ boundary under a forward-propagating hook, the effect persists ($\Delta^{\mathrm{fwd}}_{\mathrm{decode}}{=}{+}0.0082$ at L57, CI $[{+}0.0055,{+}0.0113]$) and replicates cross-family on Llama-3.3-70B (App~\ref{app:xbench}). In neither family does the agent's realised next call move ($0\%$ shift). We therefore read this as evidence that dependency information \emph{propagates} through the residual stream, not as behavioural control; behaviourally-sufficient and cross-domain causal interventions remain future work (Limitations).

\subsection{Generality of the chain-composition signal}
\label{sec:claimB}

% Results 4.2.1 Chain composition (was top-level §5 HEADLINE; demoted).
% \label{sec:chain} preserved. F3 lives here. Numbers: DATA_INVENTORY §5 ONLY.
\subsubsection{Chain composition}
\label{sec:chain}

\begin{figure}[t]
\centering
\resizebox{\columnwidth}{!}{%
\begin{tikzpicture}[
  font=\sffamily,
  >={Stealth[length=2mm]},
  call/.style={draw=cBlue,thick,rounded corners=2pt,fill=cBlue!7,
               minimum width=13mm,minimum height=8mm,align=center,font=\small},
  direct/.style={->,cBlue,thick},
  trans/.style={->,cGreen!60!black,very thick,dashed},
]
\node[call] (c0) {find\_\\user};
\node[call,right=12mm of c0] (c1) {get\_\\order};
\node[call,right=12mm of c1] (c2) {get\_\\product};
\node[call,right=12mm of c2] (c3) {modify\\order};
\node[call,right=12mm of c3] (c4) {confirm};
% direct oracle edges (solid, along the chain + one skip)
\foreach \a/\b in {c0/c1,c1/c2,c2/c3,c3/c4}{\draw[direct] (\a)--(\b);}
\draw[direct,bend right=20] (c1) to (c3);
% transitive-only edges the probe recovers (dashed, arcs above)
\draw[trans,bend left=38] (c0) to (c2);
\draw[trans,bend left=46] (c0) to (c3);
\draw[trans,bend left=52] (c0) to (c4);
\draw[trans,bend left=40] (c1) to (c4);
% legend
\node[anchor=west,font=\scriptsize] at ([yshift=-9mm]c0.south west)
  {\textcolor{cBlue}{\textbf{---\!\!\textgreater}}\, direct oracle edge
   \quad
   \textcolor{cGreen!60!black}{\textbf{- -\!\!\textgreater}}\,
   transitive-only edge recovered by the probe};
\end{tikzpicture}}
\caption{\textbf{The probe recovers the \emph{transitive closure} of the dependency DAG --- multi-hop provenance edges that are never adjacent in the trajectory.} A representative trajectory: solid blue = direct oracle edges (call $i$'s output feeds call $i{+}1$); dashed green = \emph{transitive-only} edges ($i \leadsto k$ with no direct $i{\to}k$) that the residual-stream probe linearly recovers at AUROC 0.986 --- higher than the 0.869 direct-edge AUROC (\S\ref{sec:chain}, Figure~\ref{fig:twin}).}
\label{fig:transdag}
\end{figure}
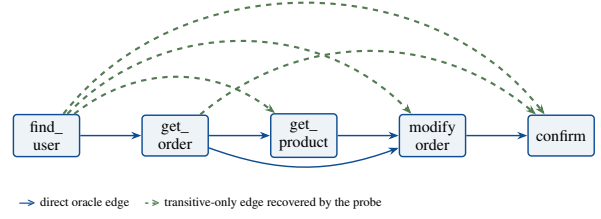

\S\ref{sec:pairwise} shows the residual stream represents \emph{pairwise} dependency edges. Whether it encodes the oracle DAG's \emph{transitive closure} --- multi-hop provenance, $i \leadsto j$ through intermediate calls --- as a jointly coherent object is a stronger question (Figure~\ref{fig:transdag}). We test it at the same probe granularity, then check robustness to seed choice and model family, and characterise where in the stack the signal emerges.

\begin{figure}[t]
\centering
\IfFileExists{figures/F3_layerwise_twin.pdf}{\includegraphics[width=\columnwidth]{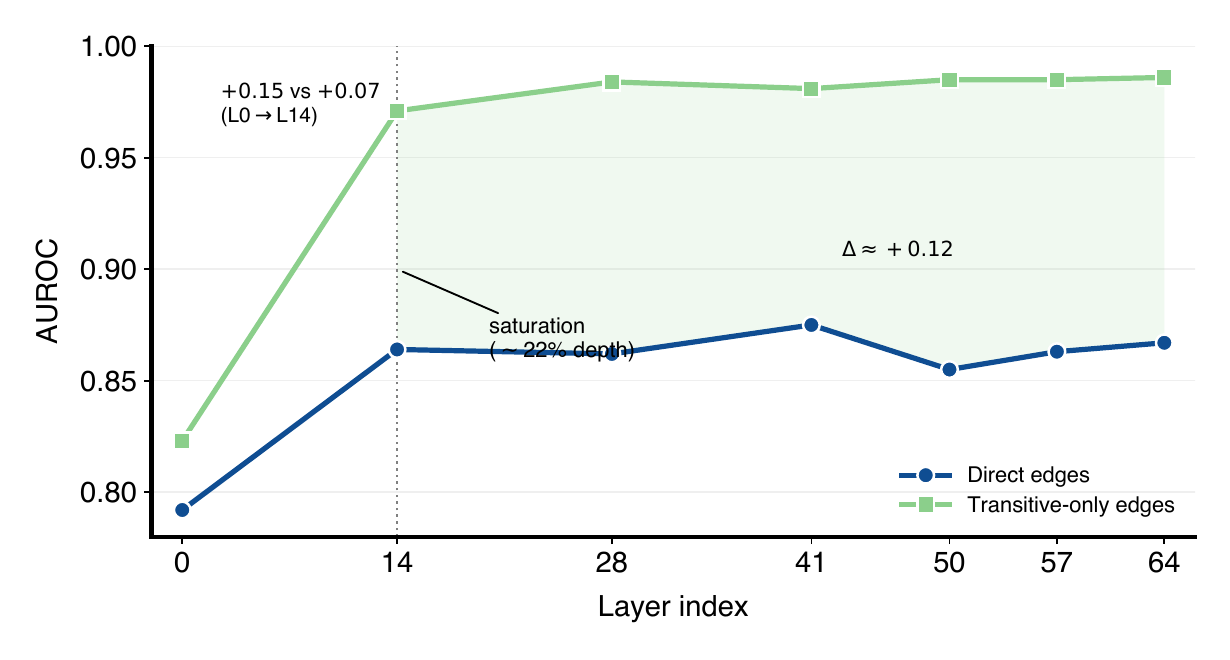}}{\fbox{\begin{minipage}{0.95\columnwidth}\centering\vspace{2cm}\textbf{[F3 pending render]}\vspace{2cm}\end{minipage}}}
\caption{\textbf{Multi-hop (transitive) provenance is decoded \emph{more} strongly than direct edges, and the two co-emerge in the early-mid stack.} Per-trajectory \textsc{logo} AUROC of the direct-edge probe (blue) and the transitive-only probe (green) across seven sampled layers of Qwen3-32B: both saturate by layer 14 ($\sim$22\% depth); the transitive signal rises $\sim$2$\times$ faster over L0--L14 ($+0.15$ vs $+0.07$ AUROC) and stays $\sim$$+0.12$ AUROC above the direct signal through layer 64 --- composition is built up \emph{alongside} direct dependency, not as a late-stack operation.}
\label{fig:twin}
\end{figure}

\textbf{The residual encodes multi-hop provenance \emph{better} than direct edges.} A transitive-only probe (label $y{=}1$ iff $i$ is a multi-hop ancestor of $j$ and $(i,j)$ is not a direct edge; 103 positives across 105 trajectories) attains AUROC \textbf{0.986} (paired-bootstrap 95\% CI $[0.977, 0.993]$; Hewitt--Liang random-label margin $+0.395$, $p<0.005$), versus the direct probe's 0.869 on the same pool ($\Delta_{\text{direct}-\text{trans}}{=}-0.117$, paired CI $[-0.189, -0.055]$, $P(\Delta{\geq}0){=}1/2000$). The model is \emph{more} linearly decodable for ``$i$ is a multi-hop ancestor of $j$'' than for ``$i$ is a direct parent of $j$'' once a comparable negative set is used. Plausibly, transitive relations reflect more global trajectory-level plan structure; the gap may also depend on the negative pool, since the transitive-only label set excludes direct edges from the positives --- the AUROC gap is therefore against a structurally easier negative set, not a stronger encoding alone.

\textbf{The non-positional contribution more than doubles over \S\ref{sec:pairwise}.} On the transitive task a 5-feature positional baseline attains AUROC 0.823; the residual-only V1 probe attains 0.986, a non-positional conditional contribution of \textbf{+0.163} (paired-bootstrap 95\% CI $[+0.106, +0.222]$, $P(\Delta{\leq}0){=}0/2000$) --- more than $2\times$ the $+0.0775$ contribution on the direct task (\S\ref{sec:pairwise}). The joint minus residual-only difference is $\approx 0$: on this task position is fully subsumed by the residual representation.

\textbf{Seed- and family-robust.} Re-running the full pipeline on three Qwen3-32B seeds gives transitive conditional contributions $+0.163$, $+0.155$, $+0.147$ --- mean \textbf{+0.155 $\pm$ 0.008} (each $P(\Delta{\leq}0){=}0/2000$; transitive AUROC $0.977 \pm 0.012$, direct $0.889 \pm 0.022$). This makes the result unlikely to be a lucky-seed artefact (SD $0.008$, $6\times$ under a pre-registered $0.05$ threshold). Cross-family, Llama-3.3-70B replicates at $+0.134$ (paired CI $[+0.093, +0.178]$, $P(\Delta{\leq}0){=}0/2000$; direct 0.912, transitive 0.967), only $+0.021$ from the Qwen3 mean, reducing the concern that the effect is Qwen3-32B-specific. We report \textbf{+0.155 $\pm$ 0.008} as the canonical chain-composition contribution; the single-seed $+0.163$ is its upper instance.

\textbf{Composition co-emerges with direct dependency, not late-stack.} Figure~\ref{fig:twin} re-runs both probes per layer. Both saturate at layer 14 ($\sim$22\% depth); from L0 to L14 the transitive signal climbs $+0.15$ AUROC versus $+0.07$ for direct, and from L14 to layer 64 it stays a consistent $\sim$$+0.12$ above direct. Chain-composition information is built up \emph{alongside} direct dependency in the early third of the stack --- the ``composition is a late-stack operation'' hypothesis is rejected here.

\textbf{The signal strengthens with hop distance and is direction-sensitive.} Stratifying the transitive pool (Appendix~\ref{app:trans}): AUROC \emph{increases} with hop count (1-hop 0.869, 2-hop 0.974, 3+-hop 0.995) and trajectory length (2--3 calls 0.942, 4--6 0.972, 7+ 0.998) --- long-range reachability is decoded \emph{more} cleanly than local edges. Applying the forward-trained probe to direction-reversed features drops AUROC from 0.869 to 0.724, so the encoding is direction-sensitive, not a symmetric co-occurrence detector.

\textbf{Thresholded predictions are not transitively coherent.} At each trajectory's F1-optimal threshold, the binarised edge predictions show transitive consistency $P(\hat{i{\to}k} \mid \hat{i{\to}j}, \hat{j{\to}k})$ of $0.307$ versus a pairwise-independence null of $0.429$ (Wilcoxon $p{=}0.999$ one-sided obs\,$>$\,null) --- below the pairwise-independence null. This is consistent with the representational claim, which does not commit to threshold-level decisional coherence: the residual \emph{represents} the multi-hop graph (the $+0.155$ contribution above), while the thresholded \emph{decisions} cluster on direct edges, which by construction do not form $i{\to}k$ triples through an intermediate $j$. The representational signal has causal support at the probe-readout level (\S\ref{sec:causal}) and, as we show next, generalises across domains (\S\ref{sec:crossdomain}).

% Results 4.2.2 Cross-domain generality (cross-domain half of old §6, demoted).
% \label{sec:crossdomain} preserved. Table 2 + F4 live here. Numbers: DATA_INVENTORY §6.2/§6.3.
\subsubsection{Cross-domain generality}
\label{sec:crossdomain}

\textbf{The cross-domain results match the qualitative prediction} (three positive interactive multi-hop replications: $\tau$-bench retail $+0.155$, ToolHop $+0.161$, ComplexFuncBench $+0.060$; predicted-zero TaskBench $-0.02$; inapplicable BFCL; Spearman $\rho{=}-0.80$ / $n{=}4$ a tendency, not a law).

If the chain-composition signal is genuinely \emph{non-positional}, it makes a concrete, testable prediction: its magnitude must scale \emph{inversely} with how well call order alone predicts dependency, and it must \emph{vanish} where position already suffices (single-shot planning) or where no transitive structure exists (direct-only tool use). We test this prediction by re-running the \S\ref{sec:setup} probe and evaluation pipeline --- identical probe, \textsc{logo} grouping, and bootstrap protocol, with benchmark-appropriate trajectory collection (Appendix~\ref{app:xbench}) --- on four further benchmarks spanning the interactive$\times$multi-hop axis (Table~\ref{tab:xbench}).

\begin{table}[t]
\centering
\caption{\textbf{The non-positional contribution behaves consistently with a non-positional account: large where call order is a weak proxy, attenuating as position-predictability rises, absent where position suffices or no transitive structure exists.} Columns: direct / transitive AUROC, position-only baseline, $\Delta$ resid$-$pos (conditional contribution), effective $N$ (distinct-task count). All powered $\Delta$: $P(\Delta{\le}0){=}0$; TaskBench n.s.; BFCL untestable (direct-only). $^\dagger$$\tau^2$-bench telecom is underpowered for the trend (transitive$+$ 9 < pre-registered 30); recorded as corroboration only. Protocols in Appendix~\ref{app:xbench}.}
\label{tab:xbench}
\footnotesize
\setlength{\tabcolsep}{4pt}
\resizebox{\columnwidth}{!}{%
\begin{tabular}{l r r r r r}
\toprule
\textbf{Benchmark} & \textbf{Dir.} & \textbf{Tr.} & \textbf{Pos.} & \textbf{$\Delta$} & \textbf{$N$} \\
\midrule
\multicolumn{6}{l}{\emph{Interactive $\wedge$ multi-hop --- contribution testable}} \\
\quad $\tau$-bench retail & 0.889 & 0.977 & 0.79 & \textbf{+.155} & 105 \\
\quad ToolHop & 0.854 & 0.966 & 0.805 & \textbf{+.161} & 994 \\
\quad ComplexFuncBench & 0.849 & 0.943 & 0.872 & \textbf{+.060} & 400 \\
\midrule
\multicolumn{6}{l}{\emph{Interactive $\wedge$ multi-hop --- corroborative, underpowered for trend}} \\
\quad $\tau^2$-bench telecom & 0.935 & --- & 0.660 & +.269 & 60$^\dagger$ \\
\midrule
\multicolumn{6}{l}{\emph{Boundary cases --- contribution not testable, as predicted}} \\
\quad TaskBench (single-shot) & 0.894 & 0.888 & 0.891 & $-$.02 & 200 \\
\quad BFCL (direct-only) & \textbf{0.960} & --- & --- & n/a & 197 \\
\bottomrule
\end{tabular}}
\end{table}

\begin{figure}[!t]
\centering
\IfFileExists{figures/F4_domain_modulated.pdf}{\includegraphics[width=\columnwidth]{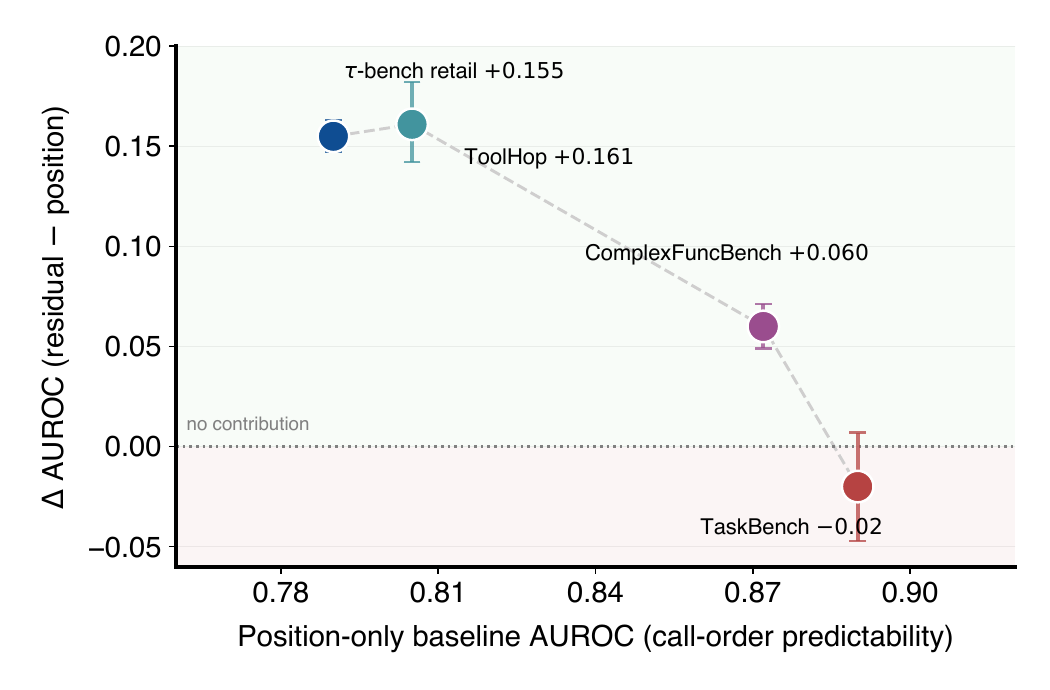}}{\fbox{\begin{minipage}{0.92\columnwidth}\centering\vspace{1.8cm}\textbf{[F4 pending render]}\\ $\Delta$ resid$-$pos vs position-predictability, 4 domains\vspace{1.8cm}\end{minipage}}}
\caption{\textbf{The non-positional contribution falls monotonically as call order becomes more position-predictable.} $\Delta$ resid$-$pos (transitive task) vs.\ each benchmark's position-only baseline; bars are 95\% paired-bootstrap CIs. Spearman $\rho{=}-0.80$ over $n{=}4$ is descriptive only ($p\approx 0.33$); the dashed line is a visual guide, not a fitted law.}
\label{fig:domains}
\end{figure}

\textbf{The qualitative pattern holds benchmark by benchmark.} Each interactive-multi-hop benchmark independently gives a strongly positive contribution with a tight paired-bootstrap CI and $P(\Delta{\leq}0){=}0$ (Table~\ref{tab:xbench}), and the single-shot control vanishes (TaskBench $-0.02$, n.s.). These per-benchmark CIs --- not a fitted trend --- are the load-bearing evidence; the four effects are additionally rank-ordered by position-predictability (Spearman $\rho{=}-0.80$, $n{=}4$), a descriptive tendency only ($p{\approx}0.33$), not a law (Figure~\ref{fig:domains}). The airline expansion is excluded (effective $N{=}13$, pre-registered $\geq 15$); $\tau^2$-bench telecom is an underpowered corroboration only (Appendix~\ref{app:xbench}).

\textbf{ComplexFuncBench shows the predicted attenuation.} Its call order is markedly more position-predictable (baseline $0.872$), so the account predicts the contribution should attenuate yet stay positive --- which $+0.060$ confirms; ToolHop ($+0.161$) instead matches retail.

\textbf{The contribution vanishes at both predicted boundaries.} The prediction also says where the contribution must \emph{disappear}: single-shot planning, where generation order already encodes dependency, and direct-only tool use, where no transitive structure exists. Both boundary cases behave as predicted. On TaskBench \citep{taskbench} --- single-shot, position-only baseline $0.891$ --- the contribution is $-0.02$ (vanishes, as predicted); and because TaskBench's dependency DAG is \emph{independently provided} (not our substring oracle), its direct AUROC $0.894$ / transitive $0.888$ provides evidence that the decodability finding is not solely an artefact of the substring oracle. On BFCL multi-turn \citep{bfcl} --- structurally direct-only (1 transitive-only edge in $4{,}525$ pairs) --- the contribution is untestable by construction, while direct AUROC \textbf{0.960} over 197 distinct stateful tasks extends decodability to a further domain.

        % §4 Results — §4.1 Claim A, §4.2 Claim B
                                   %   (04_results \inputs r1..r5)
% Conclusion — long paper §6. Numbers: ../../DATA_INVENTORY.md only.
\section{Conclusion}
\label{sec:conclusion}

We trained a low-capacity edge probe on Qwen3-32B's residual stream under Hewitt--Liang random-label and conditional-probing controls \citep{hewitt2019control, hewitt2021conditional}. The tool-call dependency graph --- pairwise and transitive --- is linearly decodable from the residual stream; per-layer activation patching adds supporting evidence that this representation propagates rather than passively reads out (behaviour not moved). The non-positional component replicates on ToolHop and ComplexFuncBench and attenuates with each benchmark's position-predictability, and a counterfactual contrast (two domains) indicates the signal tracks abstract dependency topology, not the identifier values flowing through it. More broadly, these results suggest tool-using agents expose a new class of mechanistic-interpretability targets: runtime execution structures, not only token-level decisions or reasoning traces.
     % §6 Conclusion (now §5 after merge)

% ---------- limitations (mandatory, unnumbered, not in page limit) ----------
% Limitations — long paper. Published-short-paper style: sub-headed paragraphs,
% no volunteered numerical self-attacks. Adapted+extended from FROZEN short paper.
\section*{Limitations}

\paragraph{Scope of the propagation claim.} Our patching evidence is representational propagation, not behavioural control. Activation patching shifts the downstream dependency readout (feature-level, and propagation-robust at a strictly later, non-patched call-$j$ boundary; App~\ref{app:xbench}), yet the agent's realised tool call does not move under the same intervention. This causal evidence is established on Qwen3-32B in the $\tau$-bench retail domain and replicates on Llama-3.3-70B, while decodability and the non-positional contribution are shown more broadly across benchmarks; cross-domain causal replication and behaviourally-sufficient interventions over many turns remain future work.

\paragraph{Models, scale, and agent setup.} We probe two open-weight models in the 32B--70B range; whether the same linear decodability holds at much smaller scales, or in closed frontier models, is untested. Trajectories are collected under a single agent policy and decoding configuration per benchmark, so sensitivity to the agent scaffold, prompt, and sampling settings remains uncharacterised.

\paragraph{Dependency-graph topology.} The benchmarks on which the non-positional contribution is established differ in dependency-graph shape; one corroborating benchmark (ToolHop) is composed largely of linear chains rather than branching DAGs. The contribution's behaviour on richer branching topologies, and on agent domains with error-recovery or re-planning sub-graphs, remains to be characterised.

\paragraph{Oracle construction and positional baseline.} The retail oracle uses a normalised length-$\geq 4$ substring match; an independent schema-typed value-equality oracle on the same domain (no substring matching) agrees with it at precision $1.0$ and reproduces both decodability and the activation-patching result (Appendix~\ref{app:valuecorr}), and TaskBench's independently provided ground-truth DAG (\S\ref{sec:crossdomain}) corroborates cross-domain. Both oracles define an edge by a value produced at call $i$ reappearing in call $j$'s arguments, so the construct is \emph{value-reuse} dependency: a dependency mediated by a transformed value (e.g.\ a count or reformatted field) or by pure control flow (call $j$ issued \emph{because} call $i$ returned, without reusing its value) lies outside both oracles, and ``dependency structure'' throughout should be read in this value-reuse sense. Agent-self-reported chains and full provenance instrumentation remain compatible with the same pipeline but untested. The non-positional contribution is measured relative to a 5-feature positional baseline; its magnitude depends on how strong that within-trajectory prior is in a given domain --- we report this dependence as a finding (the domain-modulation of \S\ref{sec:crossdomain}) rather than treating any single domain's magnitude as the definitive effect size.

\clearpage
\bibliography{refs}

\appendix
% Implementation Details — Appendix A. Adapted from FROZEN short-paper App A
% + long-paper extension paragraph. Numbers: ../../DATA_INVENTORY.md §2/App A.
\section{Implementation Details}
\label{app:impl}

\textbf{Model and inference.} Qwen3-32B \citep{qwen3_2025} (HuggingFace revision prefix \texttt{9216db5}) served in bf16 on one NVIDIA GH200 120\,GB GPU per worker, with a decode-time hook capturing the residual stream at each transformer block for every assistant token. Trajectory generation: \texttt{do\_sample=True}, temperature 0.6, top-$p$ 0.95, top-$k$ 20, \texttt{max\_new\_tokens}{=}4096, max turns 32. Random seed 42 across Python, NumPy, PyTorch, and scikit-learn; identical seeds across clean and corrupted runs. The $\tau$-bench commit hash is recorded with the release artefacts.

\textbf{Trajectory selection.} We collect 120 retail task instances; after filtering trajectories with $<2$ tool calls, 105 remain. The pair space comprises $1{,}129$ ordered $(i,j)$ pairs with $i<j$: 286 positives and 843 negatives.

\textbf{Probe input.} For each pair, the residual-stream vector at the token immediately following each tool call's terminating boundary, mean-pooled across layers $\{0, 14, 28, 41, 50, 57, 64\}$, then concatenated: $x_{ij} = [\bar{H}^{(\mathrm{pool})}_i \,;\, \bar{H}^{(\mathrm{pool})}_j] \in \mathbb{R}^{10{,}240}$ (V1).

\textbf{Probe training.} sklearn \texttt{LogisticRegression}, $L_2$ penalty, $C{=}0.01$, solver \texttt{lbfgs}, \texttt{max\_iter}{=}2000, \texttt{class\_weight='balanced'}; \texttt{StandardScaler} fit on the training fold only.

\textbf{Cross-validation and CI.} Per-trajectory \texttt{LeaveOneGroupOut} (105 folds on retail); AUROC computed once over concatenated out-of-fold predictions. Bias-corrected-and-accelerated bootstrap, 2{,}000 trajectory-level resamples. Hewitt--Liang control: 500 label permutations for the main probe (200 for downstream variants), probe refit from scratch each time; empirical $p$ is the fraction of permutations with AUROC $\geq$ observed.

\textbf{Input ablations.} V0--V4 vary which boundary tokens feed the probe and the number of pooled layers; V1 is canonical (Table~\ref{tab:main}).

\textbf{Oracle construction (full specification).} For an ordered call pair $(i,j)$, $i<j$, let $o_i$ be call $i$'s tool output rendered as text and $a_j$ the JSON serialisation (\texttt{json.dumps}, default settings) of call $j$'s arguments. Both are normalised by $s \mapsto$ \texttt{re.sub(r"{\textbackslash}s+"," ",s).strip()} (collapse every maximal whitespace run to one space; trim ends). We then enumerate every contiguous character substring of the normalised $o_i$ of length $\geq 4$, keep those that occur verbatim (case-sensitively) as a substring of the normalised $a_j$, and discard any that is contained in a longer retained hit; an edge $i\!\to\!j$ exists iff at least one maximal hit remains. No lowercasing, stemming, tokenisation, punctuation removal, or Unicode normalisation beyond whitespace is applied. The transitive-only label set is the transitive closure of these direct edges minus the direct edges. For the causal test (Appendix~\ref{app:xbench}), the 80 minimal pairs are all trajectory pairs $(A,B)$ whose tool-name sequences share a common prefix and whose oracle DAGs differ in exactly one edge on that shared prefix; the differing edge defines the patched source call. Oracle-construction and minimal-pair-selection code are released with the probe artefacts.

\textbf{Extended experiments.} The transitive (\S\ref{sec:chain}) and cross-domain (\S\ref{sec:crossdomain}) experiments reuse this pipeline unchanged. Multi-seed adds Qwen3-32B seeds 43 and 44; cross-family uses the \texttt{unsloth/Llama-3.3-70B-Instruct} mirror (ungated, identical weights) with a matched fractional-depth layer set. Activation patching (Appendix~\ref{app:xbench}) is feature-level and all-cached (no additional model inference). Cross-benchmark runs group all \textsc{logo} folds and bootstrap resamples by task identity and report the distinct-task effective $N$. Total compute is $\approx$300--340 GPU-hours of an 2{,}800-hour budget (Python 3.10.20, \texttt{kg\_verl} environment); the two cross-domain benchmarks together cost $\approx$12--15 GPU-hours.

% Value-Corruption and Positive-Control Analysis — Appendix B.
% Adapted near-verbatim from FROZEN short-paper App B (number-audited).
% Labels remapped. Numbers: ../../DATA_INVENTORY.md §4 ONLY.
\section{Value-Corruption and Positive-Control Analysis}
\label{app:valuecorr}

\textbf{Counterfactual construction.} For each of 120 source trajectories we identify the median-index tool call and swap a single \texttt{\_id} field in its output with another valid id drawn uniformly from the same database table, holding all preceding state fixed. The corrupted trajectory is replayed from the swap point and the residual stream is recaptured.

\textbf{Validation and effective propagation.} The corrupted output remains schema-valid in all cases. The corruption is oracle-aware: we walk the corrupted call's JSON for \texttt{\_id}-suffixed keys, select the field whose value appears in the oracle's downstream args, and flip 2--3 alphanumeric characters at that path. Of 120 source trajectories, $32$ (26.7\%) register an oracle-target hit; on the rest the agent's plan does not chain through the corrupted field or recovers via a fallback call. Effective $N$ for detecting propagation is therefore $\leq 32$; at this effective $N$ the paired Wilcoxon at $\alpha{=}0.05$/80\% power detects $|d|\geq 0.50$, so the observed $d{=}0.06$ is uninformative about small effects while the contrast against the structural-control $d{=}0.87$ is well outside the type-II zone.

\textbf{Matched subset.} Of 120 source trajectories, 103 (85.8\%) have $\geq 2$ tool calls in both conditions; these form the $N{=}103$ paired sample.

\textbf{Reward audit.} Clean trajectories pass $\tau$-bench's binary reward at $12.5\%$ (15/120); corrupted at $18.3\%$ (22/120). The $+5.8$\,pp difference is not significant (Fisher exact $p \approx 0.27$); we report it for transparency and do not treat the direction as evidence of anything.

\begin{figure}[t]
\centering
\includegraphics[width=\columnwidth]{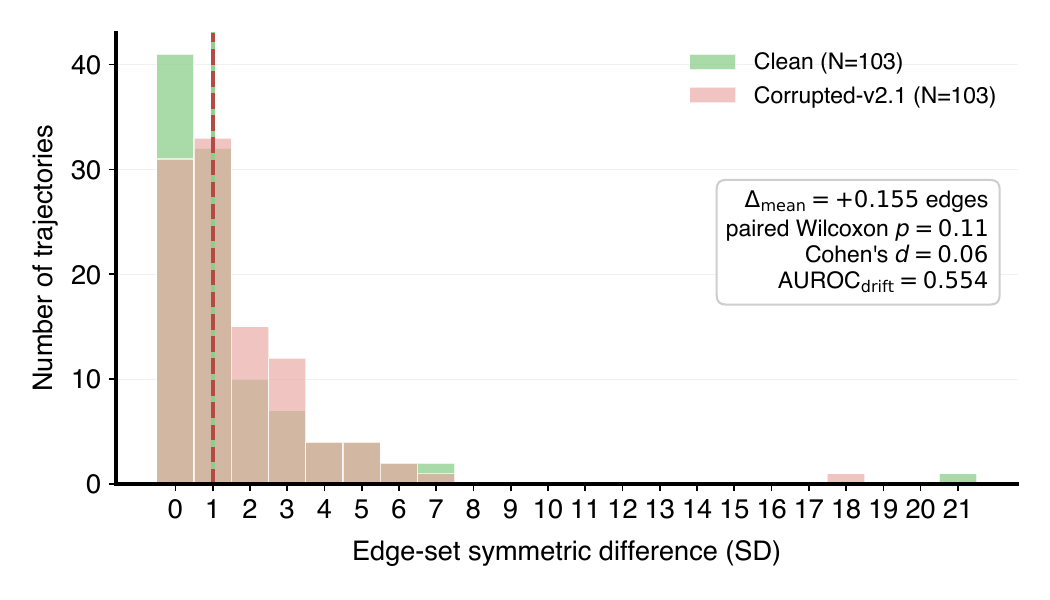}
\caption{Per-trajectory edge-set symmetric difference (SD) between probe-decoded and projected-oracle DAGs on the agent's pair space, clean vs corrupted ($N{=}103$ paired trajectories). Both distributions have median SD $= 1.0$; statistical detail in \S\ref{sec:dissoc} and this appendix.}
\label{fig:appb}
\end{figure}

\textbf{Statistics.} Paired Wilcoxon $p{=}0.11$; Cohen's $d{=}0.06$ (approximate 95\% CI $[-0.13, +0.25]$ via Borenstein paired-$d_z$ SE on $N{=}103$); mean shift $\Delta{=}{+}0.155$ edges; per-trajectory drift AUROC $0.554$. Figure~\ref{fig:appb} overlays the clean and corrupted SD distributions; medians coincide at $\mathrm{SD}{=}1.0$.

\textbf{Structural skip-tool positive control.} For each of 105 clean trajectories with $\geq 2$ tool calls we identify the median agent call index $m=n_{\mathrm{agent}}/2$ and replay with the tool response at $m$ rewritten to an empty result (literal \texttt{"\{\}"}); predictions were pre-registered (SHA-1 sealed before first inference). \emph{Per-condition decoded-vs-oracle SD:} median 1.0 in both, Cohen's $d=0.00$, drift AUROC 0.545 --- null in expectation, because the task and its oracle DAG are unchanged by a response rewrite. \emph{Plan-shift SD} between clean and corrupted decoded DAGs on the intersected pair space: median 1.0, Wilcoxon $p=1.18\times10^{-11}$, Cohen's $d=0.87$, $n=103$, 53\% non-zero shift. Together these support the \S\ref{sec:dissoc} interpretation: the probe is responsive to the structural perturbation but the adapted plan still solves the same task.

\textbf{Independent (non-substring) oracle replication.} To test whether decodability reflects abstract dependency rather than the substring rule itself, we re-derive the retail edges with a schema-typed \emph{value-equality} oracle: an edge $i\!\to\!j$ exists only where a typed-ID value produced in call $i$'s output (structured \texttt{\_id} fields and bare entity returns such as \texttt{find\_user\_id}~$\to$~\texttt{"yusuf\_rossi\_9620"}) is passed under a typed-ID key to call $j$, with no substring containment. This oracle agrees with the substring oracle at \textbf{precision $1.000$} (every independent edge is also a substring edge), recall $0.754$, F1 $0.860$, overall agreement $0.910$ over the $1{,}129$ pairs; it is a strict semantic subset ($312$ vs $414$ edges) that excludes exactly the $102$ substring-only edges --- the partial-overlap cases most likely to be lexical artefacts. Under these independent labels (per-trajectory \textsc{logo}, 2000-iteration bootstrap), the conditional contribution over the positional baseline \emph{rises} to $+0.289$ (95\% CI $[+0.265, +0.315]$, $P(\Delta{\leq}0){=}0$); the positional baseline itself falls to $0.658$, as call order predicts genuine typed-field dataflow even less well than it predicts substring overlap. Re-keying the feature-level activation patching (\S\ref{sec:causal}) to the independent oracle reproduces the causal signature: signed $\Delta_{\mathrm{patch}}>0$ with bootstrap CI excluding zero at four mid/late layers (L41 $+0.001$, L50 $+0.020$, L57 $+0.245$, L64 $+0.006$), embedding layer L0 exactly zero. Both decodability and the propagation result therefore survive a dependency oracle that makes no use of substring matching.

\begin{figure}[t]
\centering
\includegraphics[width=\columnwidth]{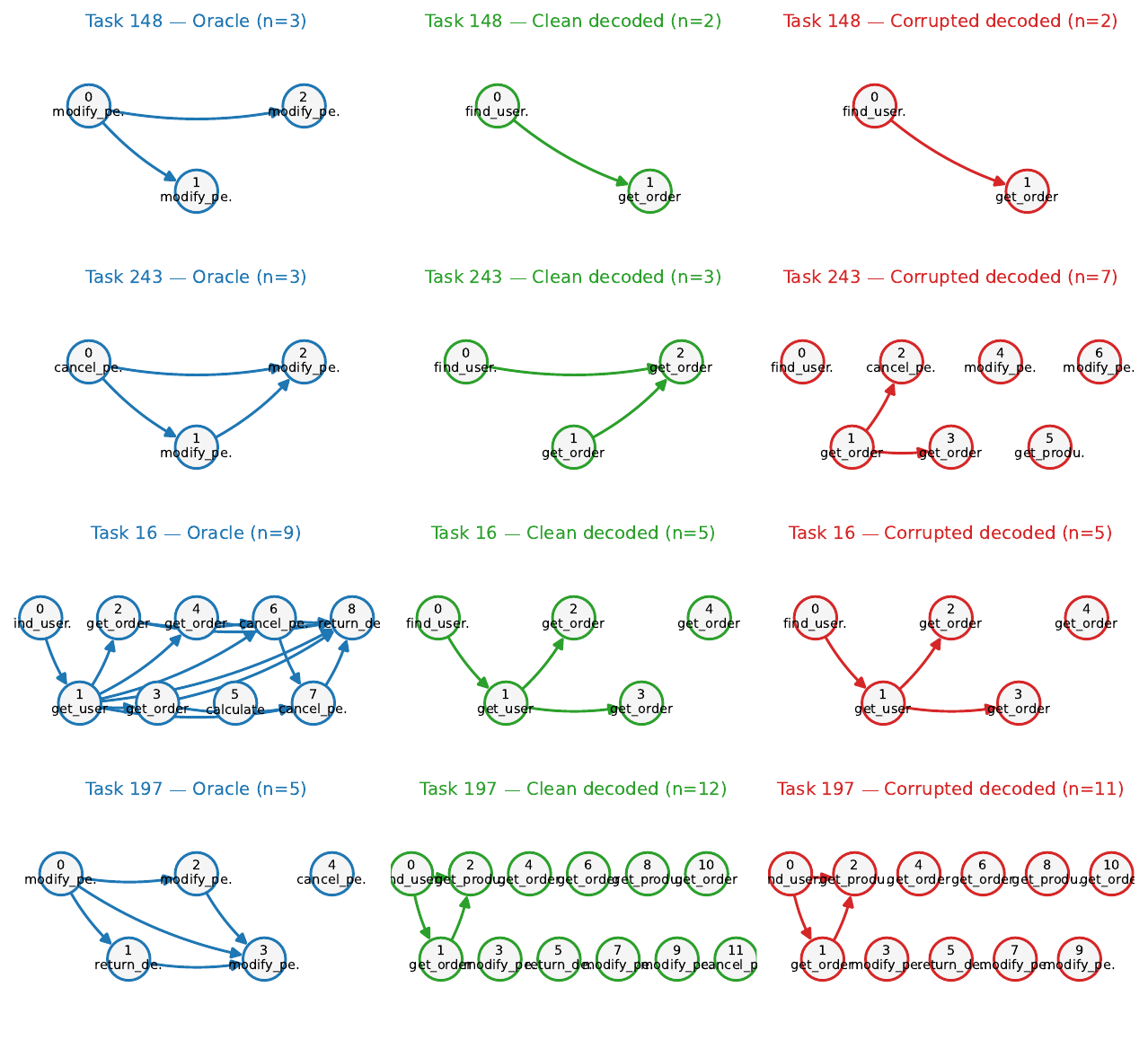}
\caption{\textbf{Probe-decoded DAGs match the oracle on representative trajectories (median edge-set symmetric difference 1), and are unchanged by value corruption.} Decoded vs projected-oracle direct-dependency DAGs for four trajectories ($\tau$-bench task IDs 148/243/16/197), spanning SD~$\in\{0,1,2\}$ and $n_{\text{agent}}\in\{2,3,5,12\}$. Columns: full oracle DAG (left, blue), clean decoded DAG (centre, green), corrupted decoded DAG (right, red); node labels are 0-indexed call positions with truncated tool names. SD is on the agent's pair space with the oracle projected to $n_{\text{agent}}$ (\S\ref{sec:dissoc}). The transitive-closure structure is shown in Figure~\ref{fig:transdag}.}
\label{fig:dags}
\end{figure}

% Random-Init and Surface-Form Controls — Appendix C.
% Adapted from FROZEN short-paper "Probe-capacity and random-init controls"
% + surface-form (T9.5) detail. Labels remapped. Numbers: DATA_INVENTORY §3.5/§3.6.
\section{Random-Init and Surface-Form Controls}
\label{app:controls}

This appendix details the conditional-probing analysis reported in \S\ref{sec:pairwise} (Table~\ref{tab:main}, ``Conditional probing'' block).

\begin{figure}[t]
\centering
\includegraphics[width=\columnwidth]{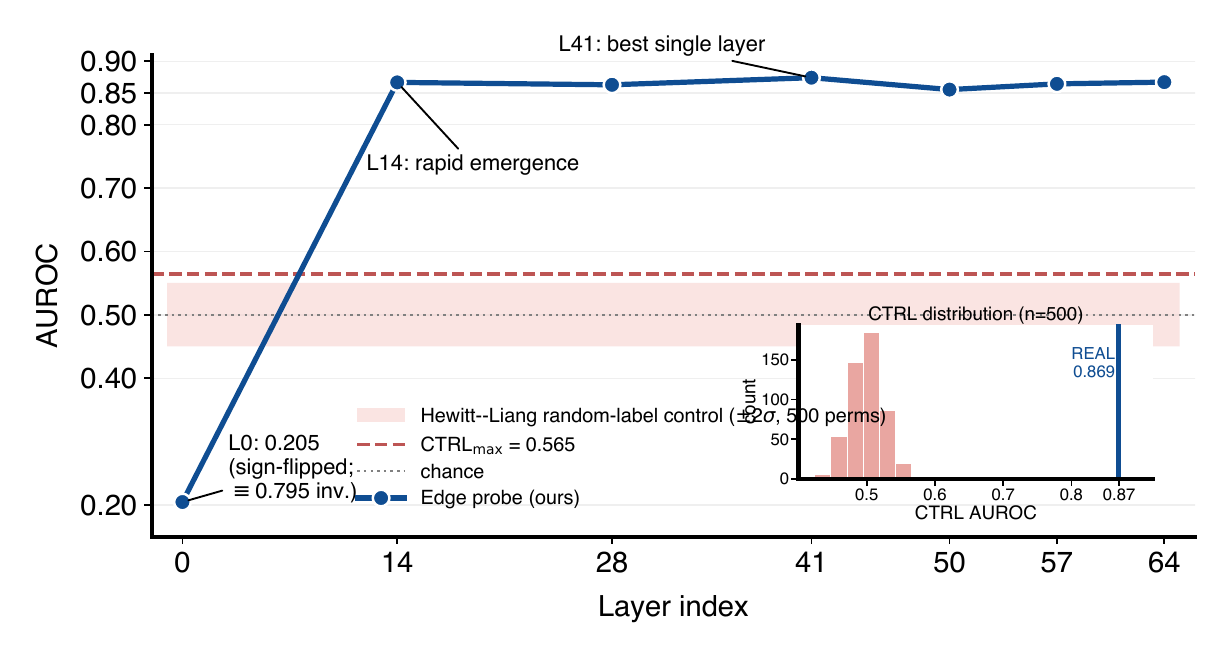}
\caption{\textbf{Per-layer edge-probe AUROC: the dependency signal emerges by layer 14 ($\sim$22\% depth) and persists to the final layer, far above the random-label control.} Qwen3-32B (blue) against the Hewitt--Liang random-label control band ($\pm 2\sigma$, light red) and the strongest of 500 permutations CTRL$_{\max}{=}0.565$ (dashed red); the embedding layer is sign-flipped under \textsc{logo} cross-validation. Inset: 500-perm control distribution with the observed multi-layer AUROC marked. Summarised in Figure~\ref{fig:overview} (bottom) and \S\ref{sec:pairwise}.}
\label{fig:layers}
\end{figure}

\textbf{Conditional-probing protocol.} Following \citet{hewitt2021conditional}: train one probe on a positional baseline alone, one on positional features concatenated with residual-stream features, and report the AUROC gap. The positional baseline is $B=[i, j, j{-}i, n_{\mathrm{agent}}, j/n_{\mathrm{agent}}]$ (5-dim, no model); the residual features are $R=[\bar{H}_i,\bar{H}_j]$ (10240-dim, V1). Same per-trajectory \textsc{logo} CV, a 200-permutation Hewitt--Liang control on the joint probe, and a 2000-iteration trajectory-level paired bootstrap on the contribution.

\textbf{Positional baseline.} The 5-feature probe attains AUROC 0.792 --- a quantification of $\tau$-bench retail's stereotyped \texttt{find\_user}$\to$\texttt{get\_order}$\to$\texttt{modify}/\texttt{cancel} layout.

\textbf{Conditional probe (trained Qwen3-32B).} AUROC 0.869, joint CI $[0.801, 0.930]$; 200-perm Hewitt--Liang CTRL$_{\max}{=}0.572$ ($0/200$ exceed observed). The headline statistic is the paired-bootstrap contribution: $\mathbf{+0.0775}$ AUROC, paired 95\% CI $[+0.032, +0.127]$, $P(\Delta{\leq}0){=}0/2000$ ($p<5{\times}10^{-4}$).

\textbf{Conditional probe (random-init Qwen3-32B).} Same architecture and tokenizer, random weights, forward-passed over the same 105 trajectories at the same boundary tokens (no resampling). AUROC 0.738; contribution $-0.053$ over the positional baseline, paired CI $[-0.101, -0.007]$ \citep{heap2025randominit}. The two paired CIs do not overlap; the training-attributable, position-controlled gap is $+0.131$ AUROC.

\textbf{Surface-form decoder.} A non-residual decoder on $n$-gram overlap between call outputs and downstream args, tool-name one-hots, and the 5 positional scalars (36-dim) attains AUROC 0.830 (paired-bootstrap CI $[0.748, 0.898]$; predictions pre-registered before inference). Residual minus surface is $\mathbf{+0.039}$ (paired CI $[+0.012, +0.063]$, $P(\Delta{\leq}0){=}0.0025$); adding surface features on top of the residual probe gains only $+0.001$, so the residual representation predictively subsumes the tested surface feature set (representational subsumption vs.\ a regularisation ceiling is not separately established).

\textbf{MLP probe.} A 2-hidden-layer head ($10{,}240 \to 1024 \to 256 \to 1$, ReLU, dropout 0.2, Adam $lr{=}10^{-4}$, early-stop) on V1 features attains AUROC 0.866 --- no improvement over the linear probe (0.869), consistent with the structural signal being linearly readable.

\textbf{Random-init diagnostics.} Before adding the positional baseline the random-init multi-layer probe attained 0.755. (i) All-65-layer pool: 0.733. (ii) Position-only: 0.792 (above all random-init variants). (iii) L0-only random-init: 0.533 (at chance --- the random-init signal is RoPE-propagated through depth, not embedding-localised). (iv) 200-perm Hewitt--Liang control on random-init: margin $+0.159$, $p<0.005$.

% Transitive-Probe Details and Stratifications — Appendix D (NEW).
% Numbers: ../../DATA_INVENTORY.md §5.2/§5.4/§5.5/§5.6/§5.7 + §7 ONLY.
\section{Transitive-Probe Details and Stratifications}
\label{app:trans}

\textbf{Transitive conditional decomposition (single seed).} On the transitive-only label set (103 positives / 105 trajectories): position-only AUROC 0.823, residual-only V1 0.986, joint 0.986; residual$-$position $+0.163$ (paired 95\% CI $[+0.106, +0.222]$, $P(\Delta{\leq}0){=}0/2000$); joint$-$residual $\approx 0$.

\textbf{Multi-seed.} Three Qwen3-32B seeds, full pipeline re-run:

\begin{table}[ht]
\centering
\caption{Multi-seed transitive-task results: three Qwen3-32B seeds, full pipeline re-run.}
\label{tab:t14_multiseed}
\setlength{\tabcolsep}{4pt}
\resizebox{\columnwidth}{!}{%
\begin{tabular}{l c c c}
\toprule
Seed & trans AUROC & pos-only & $\Delta$ resid$-$pos \\
\midrule
42 & 0.986 & 0.823 & $+0.163$ \\
43 & 0.964 & 0.808 & $+0.155$ \\
44 & 0.982 & 0.835 & $+0.147$ \\
\midrule
mean$\pm$SD & $0.977\pm0.012$ & --- & $\mathbf{+0.155\pm0.008}$ \\
\bottomrule
\end{tabular}}
\end{table}
Direct AUROC, 3-seed mean $0.889 \pm 0.022$; each seed $P(\Delta{\leq}0){=}0/2000$.

\textbf{Cross-family (Llama-3.3-70B).} Direct 0.912, transitive 0.967, position-only (trans) 0.832, residual-only 0.967; $\Delta$ resid$-$pos $+0.134$ (paired 95\% CI $[+0.093, +0.178]$, $P(\Delta{\leq}0){=}0/2000$); cross-family difference from the Qwen3 mean is $+0.021$. 103 usable trajectories; the transitive signal exceeds direct from layer 18 onward (matched fractional depth).

\textbf{Layer-wise.} Both probes per layer:

\begin{table}[ht]
\centering
\caption{Per-layer direct and transitive-only AUROC on Qwen3-32B; per-layer values are sign-resolved (AUROC${<}0.5$ flipped), so direct L0 is the embedding layer discussed in \S\ref{sec:pairwise}.}
\label{tab:t13_layerwise}
\setlength{\tabcolsep}{4pt}
\resizebox{\columnwidth}{!}{%
\begin{tabular}{l c c c c c c c}
\toprule
Layer & 0 & 14 & 28 & 41 & 50 & 57 & 64 \\
\midrule
direct & 0.795 & 0.864 & 0.862 & 0.875 & 0.855 & 0.863 & 0.867 \\
trans  & 0.823 & 0.971 & 0.984 & 0.981 & 0.985 & 0.985 & 0.986 \\
\bottomrule
\end{tabular}}
\end{table}
Both saturate by layer 14; transitive rises $+0.15$ vs direct $+0.07$ over L0--L14 and stays $\sim$$+0.12$ above direct through L64.

\textbf{Stratifications.} Multi-hop: 1-hop 0.869, 2-hop 0.974, 3+-hop 0.995. Length: 2--3 calls 0.942, 4--6 0.972, 7+ 0.998. Tool-pair heterogeneity: top \texttt{find\_user\_id}$\to$\texttt{get\_product} 1.000, bottom \texttt{get\_user\_details}$\to$\texttt{get\_product} 0.639. Direction: forward-trained probe applied to direction-reversed features drops 0.869$\to$0.724.

\textbf{Downstream-utility per-horizon.} Early-failure prediction by horizon $K$ (edge vs best positional baseline B-pos), class balance 13 pass / 92 fail; headline $K{=}2$: edge 0.616, B-pos 0.653, $\Delta_{\text{edge}-\text{pos}}{=}-0.037$, paired CI $[-0.11, +0.04]$, $P(\Delta{\leq}0){=}0.83$. Edge vs per-call at $K{=}2$: edge 0.616, per-call 0.675, $\Delta{=}-0.059$, paired CI $[-0.16, +0.04]$, $P(\Delta{\leq}0){=}0.87$; standalone per-call probe AUROC 0.884 on 481 calls (198 referenced / 283 not). Cross-domain transfer (retail-trained, zero-shot airline): AUROC 0.668, CI $[0.272, 0.903]$, $N{=}16$ trajectories / 14 projected positive edges.

% Cross-Benchmark Details — Appendix E (NEW).
% Numbers: ../../DATA_INVENTORY.md §6.1/§6.2/§6.3/§6.4 ONLY.
\section{Cross-Benchmark Details}
\label{app:xbench}

\textbf{Activation patching.} 80 minimal pairs of trajectories sharing a tool-name prefix but differing in one oracle edge; 1{,}750 patch observations. For each minimal pair $(A,B)$ and each differing pair $(i,j)$, layer $L$ of $B$'s call-$i$ residual is replaced with $A$'s layer-$L$ call-$i$ residual, the pooled probe feature is recomputed, and the signed shift toward $A$'s oracle is measured. Estimator is feature-level mediation, all-cached.

\begin{table}[ht]
\centering
\caption{Feature-level activation patching on Qwen3-32B: per-layer signed $\Delta_{\mathrm{patch}}$.}
\label{tab:t28_qwen}
\setlength{\tabcolsep}{4pt}
\resizebox{\columnwidth}{!}{%
\begin{tabular}{l c c c}
\toprule
Layer & mean signed $\Delta_{\mathrm{patch}}$ & CI excl.\ 0 & frac.\ $\to$ donor \\
\midrule
0  & $+0.00000$ & --- (=0) & 0\% \\
14 & $+0.00005$ & yes & 67.5\% \\
28 & $+0.00018$ & yes & 65.0\% \\
41 & $+0.00022$ & yes & 75.0\% \\
50 & $+0.00151$ & yes & 93.75\% \\
57 & $+0.01605$ & yes & \textbf{97.5\%} \\
64 & $+0.00158$ & yes & 95.0\% \\
\bottomrule
\end{tabular}}
\end{table}
Signed $\Delta_{\mathrm{patch}}>0$ with bootstrap CI excluding zero at every layer $\geq 14$; the embedding layer is exactly zero (negative control). Peak absolute effect $+0.016$ at layer 57; direction consistent across the 80 pairs.

\begin{figure}[t]
\centering
\IfFileExists{figures/F5_causal_patching.pdf}{\includegraphics[width=\columnwidth]{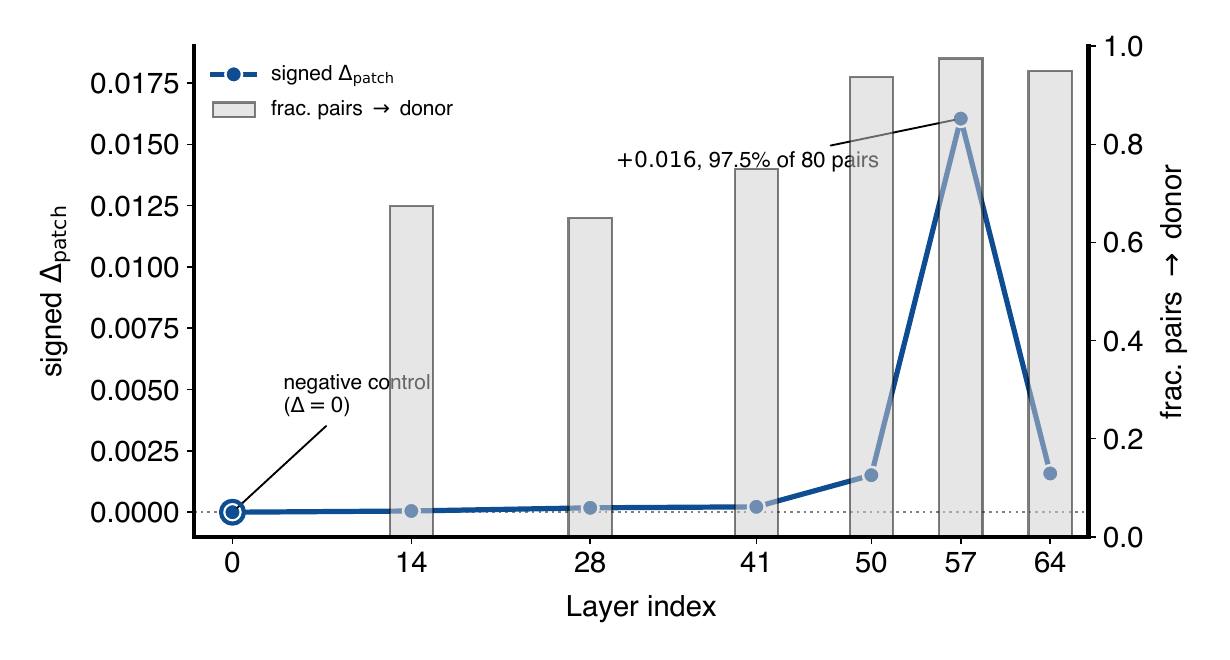}}{\fbox{\begin{minipage}{0.92\columnwidth}\centering\vspace{1.8cm}\textbf{[F5 pending render]}\vspace{1.8cm}\end{minipage}}}
\caption{\textbf{Activation patching: the dependency representation is load-bearing for the downstream readout at every layer $\geq 14$.} Per-layer signed patch effect $\Delta_{\mathrm{patch}}$ (left, blue) and the fraction of 80 minimal pairs shifting toward the donor's oracle (right, grey bars); bootstrap CI excludes zero at every layer $\geq 14$, rising to 97.5\% of pairs at layer 57, and the embedding layer (L0) is exactly zero --- a clean negative control. Summarised in \S\ref{sec:causal}.}
\label{fig:causal}
\end{figure}

\textbf{Forward-hook propagation evidence.} The feature-level estimator recomputes the pooled probe feature from the patched residual; a linear probe's output therefore moves nearly by construction when its own input is patched. We extend it with a KV-cache-safe \emph{forward-propagating} hook: the patch is applied at layer $L$ on the prefill pass and propagates to all subsequent tokens, and the probe is read at a strictly later, \emph{non-patched} call-$j$ boundary (same 80 minimal pairs).

\begin{table}[ht]
\centering
\caption{Forward-hook patching on Qwen3-32B: per-layer $\Delta^{\mathrm{fwd}}_{\mathrm{decode}}$ at a downstream non-patched call-$j$ boundary.}
\label{tab:t40_qwen}
\setlength{\tabcolsep}{4pt}
\resizebox{\columnwidth}{!}{%
\begin{tabular}{l c c c}
\toprule
Layer & mean $\Delta^{\mathrm{fwd}}_{\mathrm{decode}}$ & 95\% CI & frac.\ $\to$ donor \\
\midrule
0 (neg.\ ctrl) & $+0.0038$ & $[-0.0133, +0.0223]$ & 50.0\% \\
14 & $-0.0033$ & $[-0.0100, +0.0030]$ & 46.2\% \\
41 & $-0.0113$ & $[-0.0220, -0.0009]$ & 55.0\% \\
\textbf{57} & $\mathbf{+0.0082}$ & $\mathbf{[+0.0055,+0.0113]}$ & \textbf{87.5\%} \\
\bottomrule
\end{tabular}}
\end{table}
At L57 the propagation-robust effect on the downstream non-patched boundary is positive with CI excluding zero (87.5\% of 80 pairs); L0 (negative control) CI includes zero. L41 shows a significant \emph{negative} deflection that we report rather than hide as a layer-specific propagation artefact; the pre-registered gate requires only $\geq 1$ mid/late layer with positive CI excluding zero, satisfied by L57. A second readout records the agent's \emph{realised} next tool call under the forward-propagating patch: 0\% shift toward $A$ at both L0 and L57, $\Delta$NLL$_{\to A}\approx 0$. The forward-hook evidence is therefore \emph{representational propagation}, not behavioural control.

\textbf{Cross-family causal replication on Llama-3.3-70B.} The same 80-minimal-pair protocol replicates on Llama-3.3-70B (80 decoder layers, hidden 8192). Patch layers $\{0, 20, 60, 78\}$ chosen proportional to Qwen3's $\{0, 14, 41, 57\}$.

\begin{table}[ht]
\centering
\caption{Feature-level patching on Llama-3.3-70B (cross-family replication).}
\label{tab:t28_llama}
\setlength{\tabcolsep}{4pt}
\resizebox{\columnwidth}{!}{%
\begin{tabular}{l c c c}
\toprule
Layer & mean $\Delta_{\mathrm{patch}}$ & 95\% CI & frac.\ $\to$ donor \\
\midrule
0 (neg.\ ctrl) & $+0.00000$ & $[0, 0]$ & 0\% \\
20 & $+3.6{\times}10^{-5}$ & incl.\ 0 & 52.5\% \\
\textbf{60} & $\mathbf{+0.00285}$ & $\mathbf{[+0.00247, +0.00343]}$ & \textbf{100\%} \\
\textbf{78} & $\mathbf{+0.00991}$ & $\mathbf{[+0.00810, +0.01135]}$ & \textbf{98.75\%} \\
\bottomrule
\end{tabular}}
\end{table}

\begin{table}[ht]
\centering
\caption{Forward-hook patching on Llama-3.3-70B (cross-family replication).}
\label{tab:t40_llama}
\setlength{\tabcolsep}{4pt}
\resizebox{\columnwidth}{!}{%
\begin{tabular}{l c c c}
\toprule
Layer & mean $\Delta^{\mathrm{fwd}}_{\mathrm{decode}}$ & 95\% CI & frac.\ $\to$ donor \\
\midrule
0 (neg.\ ctrl) & $+0.0799$ & $[-0.0114, +0.1757]$ & 62.5\% \\
20 & $-0.0398$ & $[-0.0658, -0.0139]$ & 36.2\% \\
\textbf{60} & $\mathbf{+0.0838}$ & $\mathbf{[+0.0340, +0.1339]}$ & \textbf{60.0\%} \\
\textbf{78} & $\mathbf{+0.0685}$ & $\mathbf{[+0.0416, +0.0951]}$ & \textbf{63.7\%} \\
\bottomrule
\end{tabular}}
\end{table}

Feature-level cross-family STRONG: $\Delta_{\mathrm{patch}}>0$ with CI excluding zero at L60 (100\% of 80 pairs) and L78 (98.75\%); L0 control exactly zero. Forward-hook cross-family PARTIAL: representational propagation confirmed at L60 and L78 (CIs exclude zero); the L20 significant negative deflection is analogous to Qwen3 L41's. Behavioural readout matches Qwen3 ($0\%$ shift at L78, $\Delta$NLL$_{\to A}\approx 0$): behaviour does not move on Llama either. The L60/L78 forward-propagation effects are $\sim 10\times$ Qwen3's L57 magnitude, suggesting a stronger propagation-robust signal in the larger model.

\textbf{Cross-benchmark collection protocols.} All runs reuse the \S\ref{sec:setup} probe and group \textsc{logo} folds / bootstrap resamples by task identity; effective $N$ is the distinct-task count. \emph{TaskBench}: 200 single-shot planning instances; labels are TaskBench's \emph{provided} per-instance dependency DAG, not our substring oracle (1{,}504 pairs, 556 direct$+$ / 802 transitive$+$). \emph{BFCL multi\_turn\_base}: 197 distinct stateful multi-turn tasks; structurally direct-only (1 transitive-only edge in 4{,}525 pairs). \emph{ComplexFuncBench}: teacher-forced replay of the dataset's golden interactive trajectories (real args + golden observations), residual captured at each assistant decision boundary; 400 \texttt{Cross-*} multi-hop tasks, 6{,}780 pairs (2{,}747 direct$+$ / 1{,}679 transitive$+$). \emph{ToolHop}: free-generation agent loop with local-Python execution of the dataset's own functions, residual at generation step 0; 994 productive tasks, 9{,}676 pairs (2{,}938 direct$+$ / 826 transitive$+$).

\textbf{Cross-benchmark results.} Direct AUROC: TaskBench 0.894, ComplexFuncBench 0.849, ToolHop 0.854, BFCL 0.960. Transitive AUROC: TaskBench 0.888, ComplexFuncBench 0.943, ToolHop 0.966. Non-positional contribution $\Delta$ resid$-$pos (transitive task): ToolHop $+0.161$ (paired 95\% CI $[+0.142, +0.182]$, $P(\Delta{\leq}0){=}0$, position-only 0.805), ComplexFuncBench $+0.060$ (CI $[+0.049, +0.071]$, $P(\Delta{\leq}0){=}0$, position-only 0.872), TaskBench $-0.020$ (CI $[-0.047, +0.007]$, position-only 0.891 --- single-shot, position-trivial).

\textbf{Viability heuristic.} A pre-registered guard flags a benchmark position-trivial when its positional baseline exceeds 0.85. ComplexFuncBench (0.872) trips this guard yet its residual still contributes a significant $+0.060$ ($P(\Delta{\leq}0){=}0$); we report this explicitly (\S\ref{sec:crossdomain}) so the heuristic flag is not read as a negative result. The guard's intended target is the TaskBench regime, where the residual adds nothing.

\textbf{Underpowered corroborations.} A $\tau$-bench airline expansion yields effective $N{=}13$ distinct productive task structures, below the pre-stated $\geq 15$ viability threshold; we do not report it as a powered cross-domain row. A $\tau^2$-bench telecom run (free-generation Qwen3-32B with native tool calling on the Sierra $\tau^2$-bench telecom env, 60 productive distinct tasks, 282 pairs, 206 direct$+$, 9 transitive$+$) falls below the pre-registered $\geq 30$ transitive$+$ trend-inclusion threshold but yields a significant positive $\Delta$ resid$-$pos $+0.269$ (paired-bootstrap CI $[+0.099, +0.388]$, $P{=}0.0085$, position-only $0.660$), recorded as an independent interactive$\wedge$multi-hop corroboration of Claim B rather than a trend point.

\end{document}